\pdfoutput=1

\documentclass[11pt]{article}

\usepackage{times}
\usepackage{latexsym}
\usepackage[table,xcdraw]{xcolor} 
\usepackage{ACL2023}
\usepackage{amssymb}
\usepackage{multirow}
\usepackage{booktabs}
\usepackage{adjustbox}
\usepackage{tcolorbox}
\usepackage{graphicx}
\usepackage{hyperref}
\usepackage{enumitem}

\usepackage[T1]{fontenc}

\usepackage[utf8]{inputenc}

\usepackage{microtype}

\usepackage{inconsolata}
\usepackage{amsmath}

\title{On the Role of Model Prior in Real-World Inductive Reasoning}

\author{Zhuo Liu, Ding Yu, Hangfeng He\\
University of Rochester\\
{\{zhuo.liu, ding.yu, hangfeng.he\}@rochester.edu}
}

\begin{document}
\maketitle
\begin{abstract}
Large Language Models (LLMs) show impressive inductive reasoning capabilities, enabling them to generate hypotheses that could generalize effectively to new instances when guided by in-context demonstrations. However, in real-world applications, LLMs’ hypothesis generation is not solely determined by these demonstrations but is significantly shaped by task-specific model priors. Despite their critical influence, the distinct contributions of model priors versus demonstrations to hypothesis generation have been underexplored. This study bridges this gap by systematically evaluating three inductive reasoning strategies across five real-world tasks with three LLMs. Our empirical findings reveal that, hypothesis generation is primarily driven by the model’s inherent priors; removing demonstrations results in minimal loss of hypothesis quality and downstream usage. Further analysis shows the result is consistent across various label formats with different label configurations, and prior is hard to override, even under flipped labeling. These insights advance our understanding of the dynamics of hypothesis generation in LLMs and highlight the potential for better utilizing model priors in real-world inductive reasoning tasks.

\end{abstract}

\section{Introduction}

Large Language Models (LLMs) have drawn significant interests due to their performance on a diverse range of reasoning tasks \citep{kojima2022large}, such as mathematical reasoning, commonsense reasoning and symbolic reasoning. \textit{Inductive reasoning--} an important component of reasoning \citep{yang2022language,heit2000properties}, as a way to derive abstract hypothesis from limited specific observations, is widely regarded as a core aspect of human intelligence.


Existing studies primarily assess the inductive reasoning capabilities of LLMs \citep{wang2023hypothesis,qiu2023phenomenal,cheng2024inductive} by evaluating their ability to generate textual hypotheses based on in-context input-output pairs and subsequently test these hypotheses on unseen examples, thereby evaluating their generalization abilities. These studies demonstrated that LLMs can propose high-quality hypotheses, establishing them as exceptional hypothesis generators \cite{qiu2023phenomenal,cheng2024inductive,li2024mirage}.

LLMs employ various approaches to generate hypotheses depending on the nature of the task. For symbolic tasks, such as mathematical function discovery \citep{shojaee2024llm}, LLMs rely primarily on input-output mappings in demonstrations, often with minimal prior knowledge about the mathematical functions. In contrast, research by \citet{qi2023large} demonstrated that LLMs can formulate hypotheses solely from provided background information, leveraging the extensive and diverse knowledge gained during pre-training. In real-world applications, hypothesis generation tends to be data-driven , such as generating hypotheses for trending Twitter headline patterns \citep{zhou2024hypothesis}, where both prior knowledge and demonstrations are utilized. In these cases, the interaction between the model's task-specific priors and provided examples is mixed.


In empirical science, data-driven hypothesis generation serves as the foundational step toward scientific discovery \citep{majumder2024data,majumder2024discoverybench}. When employing LLMs for hypothesis generation, the goal is to uncover novel hypotheses that contribute fresh insights and ideas to the existing literature \citep{zhou2024hypothesis}. However, due to the combined influence of the model’s prior knowledge and the provided examples, the origin of generated hypotheses often remains unclear. For certain tasks, where LLMs are pre-trained on extensive knowledge bases, a strong model prior may even overshadow the potential for generating genuinely novel insights from the provided examples. This raises a critical question: \textbf{\textit{What is the role of model prior in real-world inductive reasoning?}}

To address this issue, this paper presents a systematic empirical study on real-world inductive reasoning problems, focusing on classification tasks, where hypotheses are generated to capture patterns specific to the positive class. We evaluate three representative baselines: direct input-output prompting \citep{qiu2023phenomenal}, iterative refinement with ranking \citep{qiu2023phenomenal,shojaee2024llm}, and HypoGeniC \citep{zhou2024hypothesis,liu2024literature}, across five diverse real-world tasks covering text, image, and image-text modalities. For each baseline, we conduct experiments where LLMs generate hypotheses both with and without demonstrations. The quality of the generated hypotheses is then evaluated from three perspectives: hypothesis-based classification performance, LLM-based assessments, and human evaluation.

Our experimental results reveal that, for real-world tasks where LLMs have been trained on substantial amounts of relevant data, task-specific model prior plays a dominant role in hypothesis generation. Notably, removing in-context demonstrations has minimal impact on the quality of the hypotheses. This trend holds consistently across three baselines with three LLMs: GPT-4o, Qwen2-VL and Gemini-pro, strongly suggesting that, counterintuitively, LLMs depend more on task-specific prior knowledge than on in-context demonstrations for generating hypotheses. Further analysis across various label configurations and formats supports this conclusion, indicating that model prior is often so robust that it is minimally affected by the provided examples.


\section{Related Work}

\paragraph{Inductive Reasoning with LLMs.} Primary studies on inductive reasoning mainly focus on evaluating their inductive reasoning capabilities. \citet{qiu2023phenomenal} evaluate LLMs by inducting rules from examples, demonstrated that LLMs are good hypothesis proposers. \citet{wang2023hypothesis} uses Python programs to select better hypothesis, thus improving the inductive reasoning performance. Besides these evaluations on symbolic tasks, \citet{yang2022language} propose to induce natural language rules from natural language facts while Hypotheses-to-Theories \citep{zhu2023large} learns rules from deduction. Similarly, \citet{honovich2022instruction} also show LLMs are able to infer a natural task description by provided demonstrations. Recently, some works employ LLMs to generate hypothesis that can describe the difference or shift between two distributions in different modalities, such as text \citep{zhong2022describing,zhong2023goal,singh2022iprompt}, and image \citep{dunlap2024describing,kim2024discovering}. Distinct from these studies, our work delves into understanding how LLMs perform inductive reasoning for real-world tasks, offering insights into their underlying mechanisms.

\paragraph{Hypothesis Generation with LLMs.} 
\citet{yang2023large} uses raw web corpus as observations to generate scientific hypothesis, and \citet{pham2023topicgpt} generates hypothesis to uncover latent topics in a text collection. In \citet{qi2023large}, it shows LLMs are good hypothesis proposers with only background knowledge. \citet{majumder2024data} provides initial evidence for LLMs to do data-driven discovery, where both search and verification of hypotheses may be carried out using a dataset alone. HypoGeniC \citep{zhou2024hypothesis} also uses LLMs to generate hypothesis from real-world labeled examples. \citet{si2024can} and \citet{baek2024researchagent} further explore the potential to generate hypothesis in research with LLMs to provide insights and ideas for the literature. Additionally, \citet{liu2024literature} combines theory-based generation and data-driven generation to get better hypothesis. However, these works do not clearly distinguish whether the hypotheses originate from hidden knowledge or provided examples—a distinction that is the central focus of our work.

\section{Natural Language Hypothesis Generation}
Let $\mathcal{Z} = \mathcal{D}_{P} \cup \mathcal{D}_{N}$ represent the labeled data for a real-world classification task $\mathcal{T}$, where $\mathcal{D}_{P}$ and $\mathcal{D}_{N}$ correspond to demonstrations of the positive ($P$) and negative ($N$) classes, respectively. Each sample in $\mathcal{Z}$ is a pair $(x, y)$, where $x$ denotes the example and $y \in \{P, N\}$ represents the label. A valid natural language hypothesis $h$, as introduced by \citet{zhong2022describing}, is expressed as a natural language string. For any example $x$, $h$ is capable of determining whether $x$ belongs to the positive or negative class.

\begin{figure}[t]
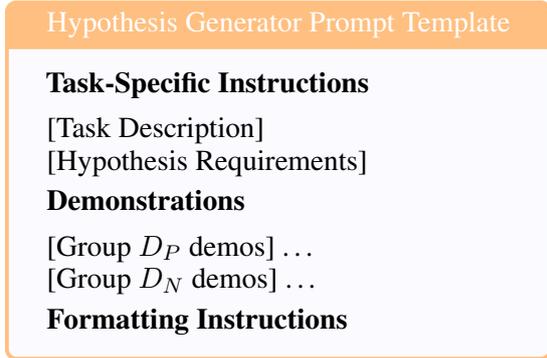

\centering
\begin{tcolorbox}[colback=blue!2!white, colframe=orange!50!white, title=Hypothesis Generator Prompt Template, coltitle=white, width= 0.45\textwidth]
\textbf{Task-Specific Instructions}\\[-10pt]  

[Task Description]\\[-15pt]  

[Hypothesis Requirements]\\[-15pt]

\vspace{3pt}  
\textbf{Demonstrations}\\[-10pt]

[Group $D_P$ demos] \dots\\[-15pt]

[Group $D_N$ demos] \dots\\[-15pt]

\vspace{3pt}  
\textbf{Formatting Instructions}
\end{tcolorbox}
\vspace{-3mm}
\caption{Prompt template for hypothesis generation.}
\label{prompt_template}
\end{figure}

Natural language hypothesis generation involves prompting LLMs to produce a set of valid hypotheses $\mathcal{H} = \{h_{1}, h_{2},\dots, h_{m}\}$ using in-context demonstrations tailored to task $\mathcal{T}$. In this paper, we consider the setting where the input to LLMs can be divided into two parts, as shown in Figure \ref{prompt_template}: (1) \textbf{Task-Specific Instructions:} a set of natural language to describe the task and the requirements for the hypothesis. (2) \textbf{Demonstrations:} a set of exemplars from different groups structured in a specified way to show the patterns of each group. Ideally, we aim to prompt LLMs to generate a list of valid hypothesis to maximize the downstream task performance, by carefully selecting instructions and demonstrations. There are two factors contributing to the hypothesis generation: \\
\textbf{Task-Specific Model Prior:} LLMs are pretrained on a diverse set of datasets, allowing them to accumulate extensive background knowledge across a wide range of domains. When provided with a task description, the model leverages its priors to infer relevant patterns, generating hypotheses based on this internalized knowledge.\\
\textbf{Input-Label Mappings in Demonstrations:} The demonstrations provided serve as a specific guidance, offering cues about how to approach the task. The model may use these demonstrations to refine its hypothesis generation, aligning its output more closely with the intended task requirements. 

\section{Experimental Settings} 

\subsection{Hypothesis Generation Baselines}

\begin{table*}[t]
    \centering
    \resizebox{0.78\textwidth}{!}{
    \begin{tabular}{llcccccccc}
    \toprule
    \multirow{2}{*}{\textbf{Dataset}} & \multirow{2}{*}{\textbf{Demos}} & \multicolumn{2}{c}{\textbf{IO-Prompting}} & \multicolumn{2}{c}{\textbf{Iterative-Refinement}} & \multicolumn{2}{c}{\textbf{HypoGeniC}} \\
    \cmidrule(lr){3-4} \cmidrule(lr){5-6} \cmidrule(lr){7-8}
      &   & \textbf{Best} & \textbf{Average} & \textbf{Best} & \textbf{Average} & \textbf{Best} & \textbf{Average} \\
     
    \midrule
    \multirow{2}{*}{\textbf{Hallucination}} & w/o & 63.7 \small{$\pm$ 2.3}  & 59.4 \small{$\pm$ 1.1}  & 66.9 \small{$\pm$ 0.5}  &62.1 \small{$\pm$ 0.3}   & 61.7 \small{$\pm$ 0.3}  & 57.9 \small{$\pm$ 0.5}  \\& w/ & 63.8 \small{$\pm$ 0.3}  & 58.3 \small{$\pm$ 0.1} & 63.7 \small{$\pm$ 2.0}  & 59.7 \small{$\pm$ 2.6} & 65.6 \small{$\pm$ 2.0}   & 59.2 \small{$\pm$ 1.1}
                                     \\
    \multirow{2}{*}{\textbf{Unhealthy Comments}}   & w/o & 70.3 \small{$\pm$ 0.7} & 63.2 \small{$\pm$ 0.9}  & 71.4 \small{$\pm$ 1.2}  & 68.8 \small{$\pm$ 1.4}  & 71.2 \small{$\pm$ 0.4}   & 67.3 \small{$\pm$ 1.3}\\   & w/  &70.0 \small{$\pm$ 0.3}  & 66.9 \small{$\pm$ 1.2} & 71.8 \small{$\pm$ 0.7} & 69.8 \small{$\pm$ 0.3} &  71.1 \small{$\pm$ 1.2}  & 67.6 \small{$\pm$ 0.8}  \\
                                   
    \multirow{2}{*}{\textbf{Funny Reddit}}  & w/o & 64.1 \small{$\pm$ 2.3} & 58.6 \small{$\pm$ 0.3}   & 67.0 \small{$\pm$ 1.6}   & 63.5 \small{$\pm$ 0.4}  & 64.4 \small{$\pm$ 1.7}  & 60.6 \small{$\pm$ 1.3} \\& w/  & 65.8 \small{$\pm$ 2.4} &  59.0 \small{$\pm$ 1.4} & 69.8 \small{$\pm$ 1.7}  & 66.1 \small{$\pm$ 0.8}  & 62.2 \small{$\pm$ 3.5} & 57.6 \small{$\pm$ 1.0} \\
                                   
    \multirow{2}{*}{\textbf{Truthful Review}} & w/o & 69.1 \small{$\pm$ 0.6}   & 57.0 \small{$\pm$ 0.5}  & 69.0 \small{$\pm$ 0.7}  & 63.8 \small{$\pm$ 1.0} & 69.2 \small{$\pm$ 0.7}  & 59.6 \small{$\pm$ 1.3}  \\ & w/  & 68.5 \small{$\pm$ 0.9}  &  59.7 \small{$\pm$ 0.8} & 69.5 \small{$\pm$ 1.6}  & 63.6 \small{$\pm$ 0.4}  & 62.4 \small{$\pm$ 5.1}  & 59.4 \small{$\pm$ 3.7} \\
                                   
    \multirow{2}{*}{\textbf{PneumoniaMNIST}}     & w/o & 75.9 \small{$\pm$ 0.5} & 72.4 \small{$\pm$ 0.4}   & 77.6 \small{$\pm$ 0.5}   & 75.6 \small{$\pm$ 0.2} & 76.8 \small{$\pm$ 0.8}  & 73.4 \small{$\pm$ 0.3}  \\    & w/  & 74.7 \small{$\pm$ 1.1} & 69.7 \small{$\pm$ 0.5}   & 76.2 \small{$\pm$ 1.7} & 74.2 \small{$\pm$ 1.1} & 74.6 \small{$\pm$ 0.5}  & 71.4 \small{$\pm$ 0.8} \\
                                   
    \midrule

    \rowcolor{gray!20}
    & w/o & \textbf{68.62}  & 62.12 & \textbf{70.38} & \textbf{66.76} & \textbf{68.66} & \textbf{63.76} \\
    \rowcolor{gray!20}
    \multirow{-2}{*} {\textbf{Overall Average}}
    & w/  & 68.56  & \textbf{62.72}  & 70.20  & 66.68 & 67.18 & 63.04 \\
    \bottomrule
    \end{tabular}
    }
    
    \caption{Accuraccy comparison of \textit{single hypothesis-based classification} across five datasets of three baselines: accuracy (\textit{mean} ± \textit{standard deviation}) for the best single hypothesis and the average across five hypotheses, with (\textbf{w/}) and without (\textbf{w/o}) demonstrations. The better overall average between (\textbf{w/}) and (\textbf{w/o}) is highlighted in \textbf{bold}. }
    \label{method}
\end{table*}

\begin{table*}[ht]
\centering
\resizebox{0.98\textwidth}{!}{
\begin{tabular}{llcccccc}
\toprule
                  &    \textbf{Demos}            & \textbf{Hallucination} & \textbf{Unhealthy Comments} & \textbf{Funny Reddit} & \textbf{Truthful Review} & \textbf{PneumoniaMNIST} & \cellcolor{gray!20}\textbf{Overall Average} \\ \midrule
\multirow{2}{*}{\textbf{Best}} & w/o         & 60.4 \small{$\pm$ 0.0}        & 68.5 \small{$\pm$ 0.0}             & 63.6 \small{$\pm$ 0.0}       & 67.0 \small{$\pm$ 0.0}          & 65.4 \small{$\pm$ 0.0}          & \cellcolor{gray!20} \textbf{64.98}            \\
                      & w/          & 60.1 \small{$\pm$ 2.3}        & 68.0 \small{$\pm$ 0.0}             & 62.4 \small{$\pm$ 2.2}       & 66.0 \small{$\pm$ 0.4}          & 62.5 \small{$\pm$ 2.9}          & \cellcolor{gray!20} 63.80             \\ \midrule
\multirow{2}{*}{\textbf{Average}} & w/o      & 57.7 \small{$\pm$ 0.0}        & 63.0 \small{$\pm$ 0.0}             & 57.1 \small{$\pm$ 0.0}       & 55.1 \small{$\pm$ 0.0}          & 57.9 \small{$\pm$ 0.0}          & \cellcolor{gray!20} \textbf{58.16}            \\
                         & w/       & 55.4 \small{$\pm$ 1.1}        & 63.4 \small{$\pm$ 0.2}             & 58.2 \small{$\pm$ 0.8}       & 56.5 \small{$\pm$ 1.1}          & 54.7 \small{$\pm$ 2.0}          & \cellcolor{gray!20} 57.64            \\ \bottomrule
\end{tabular}
}
\caption{Accuraccy comparison of \textit{single hypothesis-based classification} with \textbf{Qwen2-VL-72B}: accuracy (\textit{mean} ± \textit{standard deviation}) for the best single hypothesis and the average across five hypotheses, with (\textbf{w/}) and without (\textbf{w/o}) demonstrations. The better overall average between (\textbf{w/}) and (\textbf{w/o}) is highlighted in \textbf{bold}.}
\label{qwen}
\end{table*}

In this paper, we evaluate three commonly-used hypothesis generation baselines.
\paragraph{Input-Output Prompting.} Input-output prompting (IO-Prompting) represents the most common approach to prompting LLMs \citep{qiu2023phenomenal}. In this standard IO-Prompting framework, we directly provide the LLMs with a set of in-context demonstrations within the prompt context. The objective is to generate $m$ hypotheses that effectively captures the patterns of positive class $P$. This approach is a single-step method, utilizing the in-context demonstrations once to guide the model’s hypothesis generation.

\paragraph{Iterative Refinement with Ranking.} Standard IO-prompting utilizes in-context demonstrations only once, potentially under utilizing their full capacity. To address this limitation, various methods have been proposed to iteratively refine hypotheses, thereby enhancing model performance \citep{wang2023hypothesis, qiu2023phenomenal, shojaee2024llm, xiao2024verbalized}. In our approach, we iteratively refine hypotheses using ranking information as a feedback signal.

The refinement process begins with an initial set of $m$ hypotheses generated via IO-prompting. At each iteration, hypotheses in the bank are ranked based on their performance on a validation set. The top-ranked $m$ hypotheses are then fed back to the model, along with in-context demonstrations, guiding it to generate hypotheses with improved performance. In cases where no demonstrations are available, only the ranked hypotheses with their accuracies are provided in the iterative refinement process. This approach thus augments data utilization by continuously leveraging feedback to generate higher-quality hypotheses.

\paragraph{Update from Mistakes: HypoGeniC.} The previous methods leverage data within one single prompt to generate hypotheses, yet using all demonstrations in a single prompt may not be optimal for performance. Therefore, we also evaluate a strategy that updates hypotheses from mistakes made by current hypothesis. We largely follow an established approach, HypoGeniC \citep{zhou2024hypothesis, liu2024literature}, which iteratively generate new hypotheses from incorrect prediction examples.

In our evaluation, we initialize the hypothesis bank using standard IO-prompting as well as the reward scores as in \citet{zhou2024hypothesis, liu2024literature}. During the update phase, if the number of incorrect examples for each group reaches a predefined number, these incorrect examples are employed to guide the generation of new hypotheses. In each update, $m$ hypotheses 
 with highest reward scores are kept in the hypothesis bank. This iterative updating approach enables the model to adapt hypotheses progressively, making better use of feedback from misclassifications. For a fair comparison, when demonstrations are absent, we update the hypothesis by iterative refinement, using reward scores for ranking.

All the implementation details are in the Appendix \ref{appendix: implementation details}.

\subsection{Evaluation of Hypothesis}
After generating a set of hypotheses $\mathcal{H} = \{h_1, h_2, \dots, h_m\}$, it is crucial to evaluate their quality to ensure that the generated hypotheses are both functional and interpretable. We perform this evaluation from three perspectives: \textbf{hypothesis-based classification}, \textbf{LLM-based evaluation} and \textbf{human evaluation}. These complementary methods allow for a robust assessment, combining quantitative performance metrics with qualitative assessments from domain experts.

\begin{table*}[ht!]
    \centering
    \resizebox{0.76\textwidth}{!}{
\begin{tabular}{lcccccc}
    \toprule
    & \multicolumn{2}{c}{\textbf{IO-Prompting}} & \multicolumn{2}{c}{\textbf{Iterative-Refinement}} & \multicolumn{2}{c}{\textbf{HypoGeniC}} \\
    \cmidrule(lr){2-3} \cmidrule(lr){4-5} \cmidrule(lr){6-7}
    \textbf{Dataset} & \textbf{w/o demos} & \textbf{w/ demos} & \textbf{w/o demos} & \textbf{w/ demos} & \textbf{w/o demos} & \textbf{w/ demos} \\
    \midrule
    \textbf{Hallucination} & 62.2 \small{$\pm$ 1.0} & 61.1 \small{$\pm$ 0.3}  & 60.1 \small{$\pm$ 4.5} & 61.1 \small{$\pm$ 1.3} & 58.6 \small{$\pm$ 4.0} & 60.1 \small{$\pm$ 0.5} \\
    \textbf{Unhealthy Comments}      & 71.5 \small{$\pm$ 0.7} & 70.9 \small{$\pm$ 0.5}  & 71.0 \small{$\pm$ 0.4} & 70.9 \small{$\pm$ 0.3} & 70.9 \small{$\pm$ 1.0} & 70.7 \small{$\pm$ 2.3} \\
    \textbf{Funny Reddit}  & 58.3 \small{$\pm$ 0.4} & 59.2 \small{$\pm$ 0.3}  & 63.9 \small{$\pm$ 2.7} & 67.3 \small{$\pm$ 1.2} & 58.8 \small{$\pm$ 0.7} & 58.4 \small{$\pm$ 0.5} \\
    \textbf{Truthful Reviews}  & 63.8 \small{$\pm$ 1.4} & 65.3 \small{$\pm$ 0.9} & 68.5 \small{$\pm$ 0.3} & 69.1 \small{$\pm$ 1.3} & 67.7 \small{$\pm$ 1.5} & 62.1 \small{$\pm$ 4.6} \\
    \textbf{PneumoniaMNIST}         & 75.8 \small{$\pm$ 0.9} & 72.2 \small{$\pm$ 1.2} & 76.0 \small{$\pm$ 2.5} & 74.1 \small{$\pm$ 1.7}  & 74.9 \small{$\pm$ 1.7} & 74.6 \small{$\pm$ 1.0} \\
    \rowcolor{gray!20} \textbf{Overall Average}         & \textbf{66.32} & 65.74 & 67.90 & \textbf{68.50} & \textbf{66.18} & 65.18 \\
    \bottomrule
\end{tabular}
}
\caption{Accuracy comparison of \textit{multiple hypotheses-based classification} across five datasets of three baselines: accuracy (\textit{mean} $\pm$ \textit{standard deviation}) with (\textbf{w/}) and without (\textbf{w/o}) demonstrations. The better overall average between (\textbf{w/}) and (\textbf{w/o}) is highlighted in \textbf{bold}.}

\label{multiple}
\end{table*}
\begin{figure*}[t]
    \centering
    \includegraphics[width=0.98\textwidth]{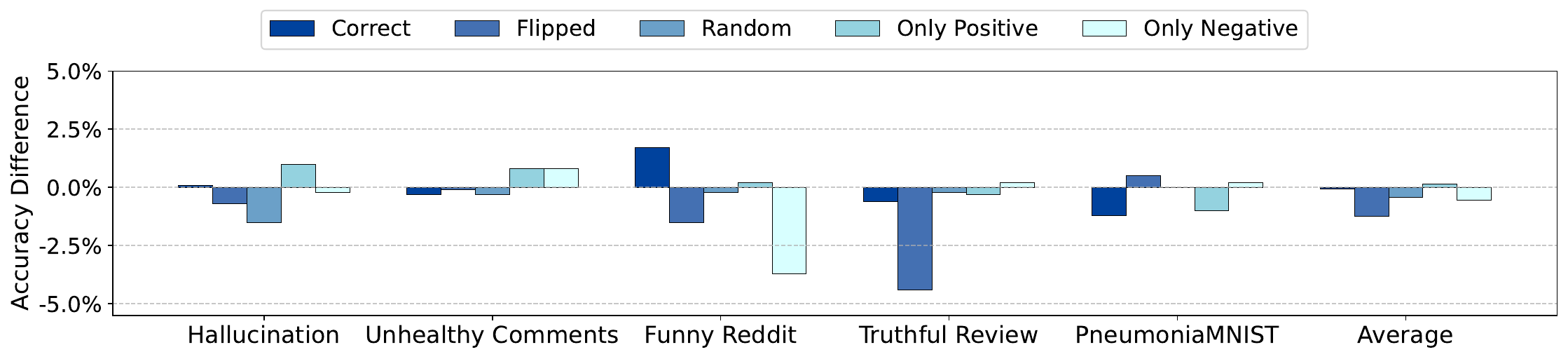}  
    \caption{Accuracy difference comparison of \textit{single hypothesis-based classification} under different label settings: Accuracy difference (\emph{accuracy of different label settings - accuracy without demos}) across five datasets with IO-Prompting. } 
    \label{fig:label_svg}
\end{figure*}
\paragraph{Hypothesis-based Inference.}
In hypothesis-based inference \citep{liu2024literature,zhou2024hypothesis}, the goal is to assess how well the generated hypotheses support downstream decision-making tasks. We measure the predictive performance of the hypothesis on a test dataset $\mathcal{D}_{\text{test}} = \{(x_j, y_j)\}_{j=1}^{N_{test}}$. The hypothesis is evaluated based on how accurately it assigns the correct label to each input $x_j$. Predictions are made by comparing test examples $ x_j $ with learned patterns, which can consist of a single hypothesis or multiple hypotheses. If a test example satisfies the pattern, it is assigned the corresponding class. Unless otherwise stated, the results reported in this work are based on patterns formed from single hypothesis. To remove the influence of prior in the inference, we also do hypothesis-based inference without knowledge, which can be found in Appendix \ref{appendix:without knowledge}. See Appendix \ref{appendix:prompt} for evaluation prompts. 
\paragraph{LLM-based Evaluation.} In addition to assessing the effectiveness of hypotheses in downstream task usage, we also evaluate their \emph{helpfulness} \citep{liu2024literature} and \emph{novelty} \citep{liu2024literature,si2024can} through LLM-based metrics. Specifically: (1) \emph{Helpfulness} measures the extent to which a hypothesis accurately captures the underlying patterns of the data and generalizes effectively to unseen samples. (2) \emph{Novelty} assesses whether the hypothesis introduces new insights or unique perspectives relevant to the task.

Our LLM-based evaluation incorporates both scoring and pairwise comparison assessments. For scoring, LLMs assign a rating on a 5-point scale to reflect each hypothesis’s quality. For pairwise comparison, we randomly pair hypotheses generated with and without demonstrations, and prompt the LLMs to select the better hypothesis in each pair. This pairwise evaluation provides insights into relative performance, while scoring offers an absolute measure of quality. 

\paragraph{Human Evaluation.} To validate the effectiveness of LLM-based evaluation, we also conduct a human evaluation to assess the quality of the generated hypotheses. Our goal is to examine the degree of alignment between LLM-based evaluation results and those obtained from human experts. Given that scoring may be challenging for human evaluators, we employ a pairwise comparison format, allowing experts to select the higher-quality hypothesis or indicate if the difference is difficult to discern. A total of nine participants are recruited for this evaluation, ensuring diverse perspectives in assessing the hypotheses.

For further details in both LLM-based and human evaluations, refer to Appendix \ref{appendix: qualitative study}.

\subsection{Other Settings}

\paragraph{Models.}
We conduct experiments with GPT-4o, Qwen2-VL-72B\footnote{https://huggingface.co/Qwen/Qwen2-VL-72B-Instruct-AWQ} and gemini-1.5-pro-002, leveraging both open-source models and API-accessible models to ensure diverse evaluation.
Unless otherwise stated, we use GPT-4o\footnote{By default, we use GPT-4o-2024-08-06. However, if a request is rejected due to safety reasons, we will switch to GPT-4o-2024-05-13.} in experiments. 

\begin{table*}[ht!]
    \centering
    \resizebox{0.96\textwidth}{!}{
    \centering
    \begin{tabular}{lcccccccc}
        \toprule
        & \multicolumn{2}{c}{\textbf{IO-Prompting}} & \multicolumn{2}{c}{\textbf{Iterative-Refinement}} & \multicolumn{2}{c}{\textbf{HypoGeniC}} & \multicolumn{2}{c}  {\cellcolor{gray!20}\textbf{Overall Average}} \\
        \cmidrule(lr){2-3} \cmidrule(lr){4-5} \cmidrule(lr){6-7} \cmidrule (lr) {8-9}
        \textbf{Criteria} & \textbf{w/o} & \textbf{w/ } & \textbf{w/o } & \textbf{w/ } & \textbf{w/o } & \textbf{w/ } & \cellcolor{gray!20} \textbf{w/o } & \cellcolor{gray!20} \textbf{w/} \\
        \midrule
        \textbf{Helpfulness} & 4.00 \small{$\pm$ 0.000} & 3.96 \small{$\pm$ 0.195} & 4.00 \small{$\pm$ 0.000} & 3.80 \small{$\pm$ 0.400} & 4.04 \small{$\pm$ 0.195} & 4.08 \small{$\pm$ 0.271}  & \cellcolor{gray!20} \textbf{4.01} & \cellcolor{gray!20} 3.95 \\
         \cmidrule(lr){2-3} \cmidrule(lr){4-5} \cmidrule(lr){6-7} \cmidrule (lr) {8-9}
        \textbf{Novelty}     & 2.56 \small{$\pm$ 0.571} & 2.40 \small{$\pm$ 0.566} & 2.60 \small{$\pm$ 0.693} & 2.60 \small{$\pm$ 0.748} & 2.84 \small{$\pm$ 0.674} & 2.36 \small{$\pm$ 0.741} & \cellcolor{gray!20} \textbf{2.67} & \cellcolor{gray!20} 2.45  \\
        \bottomrule
    \end{tabular}
    }
    \caption{LLM-based Scoring: Comparison of \emph{Helpfulness} and \emph{Novelty} scores across three baselines, with and without demonstrations (\emph{w/ demos} vs. \emph{w/o demos}). The better overall average between (\textbf{w/}) and (\textbf{w/o}) is highlighted in \textbf{bold}.}
    \label{tab:scores}
\end{table*}

\paragraph{Datasets.}
We conduct evaluations on five real-world inductive reasoning datasets: hallucination pattern induction \citep{li2023evaluating}, unhealthy comments \citep{zhong2023goal}, funny Reddit posts \citep{zhong2023goal}, pneumoniaMNIST \citep{xiao2024verbalized}, and truthful hotel reviews \citep{zhou2024hypothesis}. 

Our selection of datasets is motivated by three key factors: (1) their coverage of three distinct modalities—text (unhealthy comments, funny Reddit posts, and truthful hotel reviews), image (pneumoniaMNIST), and image-text (hallucination pattern induction), (2) diverse domains, including model behavior analysis (hallucination pattern induction), medical diagnosis (pneumoniaMNIST), and social media content (unhealthy comments, funny Reddit posts, and truthful hotel reviews), and (3) their status as widely studied problems in real-world inductive reasoning tasks. Further details and more references for these datasets are provided in Appendix \ref{appendix：dataset_split}.

\paragraph{Other Parameters.}
The number of in-context demonstrations is set to $N=30$ for IO-prompting and iterative-refinement, and $N=50$ for HypoGenic to encourage more updates. Examples are randomly sampled from the training set. For each dataset, we generate five candidate hypotheses. Main results are averaged over three random seeds to ensure robustness. More implementation details can be found in Appendix \ref{appendix: implementation details}.
\section{Task-Specific Model Prior Dominates Hypothesis Generation}
\subsection{LLMs Are Zero-Shot Hypothesis Generators}
To see the impact of the model prior in hypothesis generation, we compare the hypothesis generation in the following two settings.\\
\emph{\textbf{Model Prior Only}} is a typical zero-shot hypothesis generation scenario without the use of demonstrations, relying primarily on prior for generation. \\
\emph{\textbf{Demos with Ground Truth Labels}} is used in a typical real-world inductive reasoning tasks, with demonstrations as a specific guidance. 

Results for single hypothesis-based and multiple hypotheses-based  classification are shown in Table \ref{method} and Table \ref{multiple}. From the results, We find that removing in-context demonstrations cause little degradation for the downstream task performance. The trend is consistent across five different datasets on three baselines. In some cases, LLMs can even generate better hypothesis using only model prior. Additionally, iterative refinement outperforms the other two baselines, showing that data still helps for hypothesis selection, but not as in-context demonstrations for hypothesis generation.

\paragraph{Resutls with Qwen2-VL and Gemini-1.5-pro.}The results for single hypothesis-based classification on Qwen2-VL and Gemini-1.5-pro-002, with IO-prompting, are provided in Table \ref{qwen} and Appendix \ref{appendix:gemini}. These results similarly show a negligible performance drop without demonstrations, underscoring the universality of our findings across different models. 

\begin{figure}[t]
    \centering
    \includegraphics[width=0.49\textwidth]{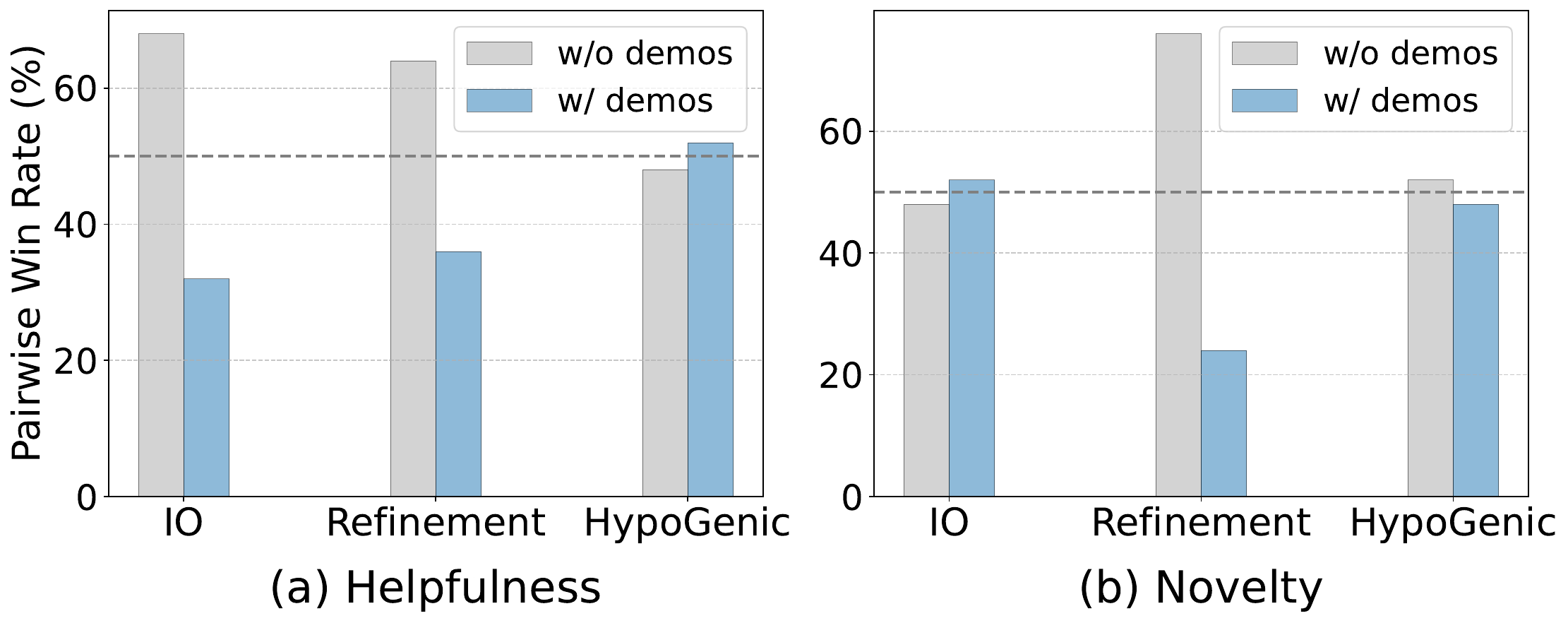}  
    \caption{LLM-based Pairwise Comparison: Pairwise win rate (\%) of three baselines. The left plot shows the comparison of \emph{Helpfulness}, while the right plot presents \emph{Novelty}. The dashed line indicates a tie where "w/ demos" and "w/o demos" perform equally well.}
    \label{llm_choice}
\end{figure}
These results indicates LLMs are good zero-shot hypothesis proposers under strong prior, and in-context demonstrations with ground truth labels are not necessary to achieve acceptable hypothesis. This is a counter-intuitive phenomenon, given that labeled data is very important in in-context learning \citep{brown2020language}, which can inform the model of corresponding data distribution \citep{min2022rethinking}.

\subsection{Input-Label Mappings in Demonstrations Cannot Override Strong Model Prior}
To further explore the interaction between model prior and input-label mappings in demonstrations in hypothesis generation, we use in-context demonstrations with different label settings: \vspace*{0.4\baselineskip}
\\ 
\hspace*{1em} (1) \emph{Demos with\textbf{ ground truth (correct) labels}}. \\
\hspace*{1em} (2) \emph{Demos with\textbf{ flipped labels}}. \\
\hspace*{1em} (3) \emph{Demos with\textbf{ random labels}}. \\
\hspace*{1em} (4) \emph{\textbf{Only positive }group demos}. \\
\hspace*{1em} (5) \emph{\textbf{Only negative }group demos}. \\
\vspace*{-0.35\baselineskip}

Figure \ref{fig:label_svg} illustrates the relative accuracy difference between various label settings and without demonstrations. From the result, there is quite limited difference (mostly smaller than $3 \%$) of performance among different settings, with the flipped label setting in truthful review as an exception, which has a performance degradation about $4.5 \%$. 

These findings suggest that while demonstrations can provide some guidance, the models' hypothesis generation abilities are ultimately shaped more by its pre-trained priors than by any superficial label configurations. Furthermore, the prior is too strong to be overridden by the patterns in demonstrations, even with totally flipped labels.

\subsection{LLM-based Evaluation Results}
\paragraph{LLM-based Scoring.}Table \ref{tab:scores} summarizes the helpfulness and novelty scores for various approaches. Each score represents the average of 25 hypotheses generated across five datasets. For helpfulness, hypotheses generated without demonstrations achieve higher scores when using IO-prompting and iterative-refinement. Regarding novelty, hypotheses generated without demonstrations score higher on IO-prompting and Hypogenic, while iterative-refinement yields a tie between the two settings.

\paragraph{LLM-based Pairwise Comparison.}Figure \ref{llm_choice} presents the pairwise comparison results for three baselines, evaluating hypotheses generated with and without demonstrations. The comparisons involve randomly paired hypotheses, with win rates aggregated across all datasets. For Helpfulness, IO prompting and iterative refinement perform better without demonstrations, while HypoGenic demonstrates improved performance with them. For Novelty, iterative refinement excels in the absence of demonstrations, whereas IO prompting and HypoGenic exhibit minimal differences between the two settings.

\begin{figure}[t]
    \centering
    \includegraphics[width=0.39\textwidth]{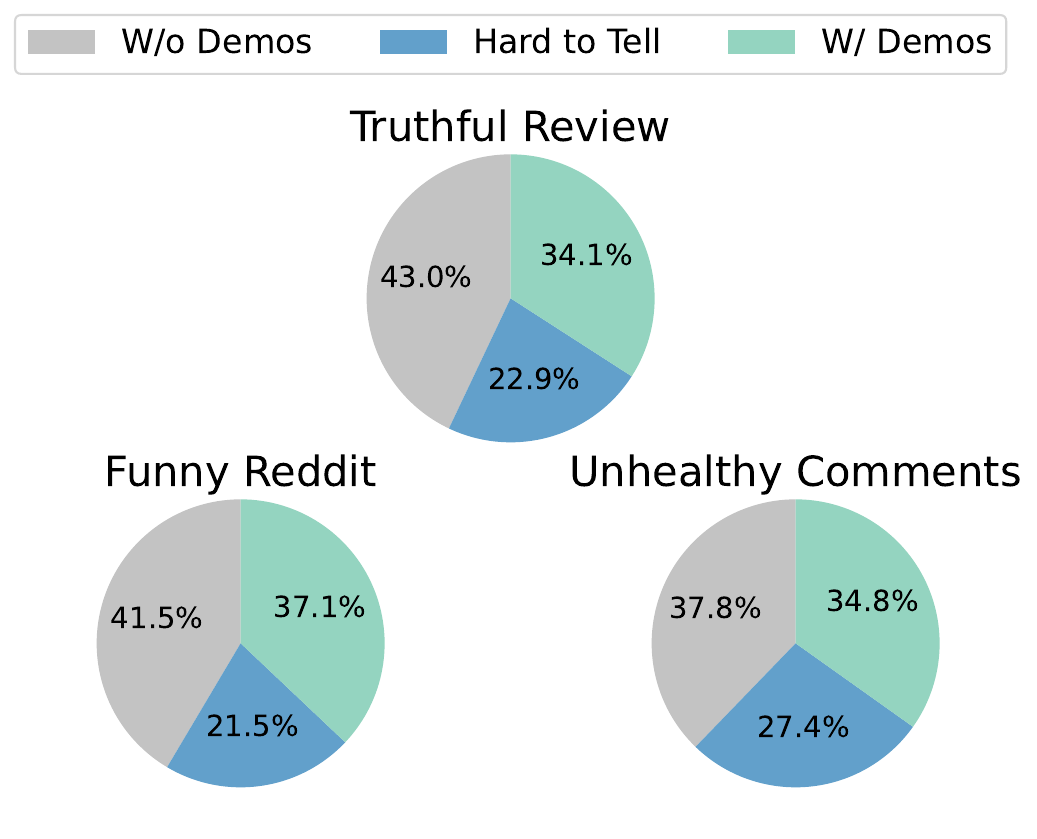}  
    \caption{Human pairwise comparison results on three datasets, showing preferences for hypotheses with, without demos, and cases where it was hard to tell the difference.}
    \label{human_choice}
\end{figure}
These results highlight that LLMs can produce highly helpful and novel hypotheses even without in-context demonstrations.

\subsection{Human Evaluation Results}
We conduct a human evaluation on Funny Reddit, Truthful Reviews, and Unhealthy Comments datasets, as the other datasets require more specialized expertise. The results are illustrated in Figure \ref{human_choice}. Across the three datasets, hypotheses generated without demonstrations received the highest percentage of preference. These findings indicate a slight overall preference for hypotheses generated using only the model's prior, though the extent of this preference varies by dataset.
\begin{table}[t]
\centering
\resizebox{0.47\textwidth}{!}{
\begin{tabular}{lcccc}
\toprule
\multirow{2}{*}{\textbf{Format}} & \multicolumn{2}{c}{\textbf{Correct Label}} & \multicolumn{2}{c}{\textbf{Flipped Label}} \\
\cmidrule(lr){2-3} \cmidrule(lr){4-5}
 & Best & Average & Best & Average \\
\midrule
\textbf{Label Format1} & 68.56 & 62.72 & 65.15 & 59.96 \\
\textbf{Label Format2} & 67.88 & 62.78 & 67.49 & 61.90 \\
\midrule
\midrule
\textbf{w/o demos} & \multicolumn{2}{c}{Best: 68.62} & \multicolumn{2}{c}{Average: 62.12} \\
\bottomrule
\end{tabular}
}
\caption{Accuracy comparison of different label formats in correct and flipped label settings with IO-prompting. Each number is the average over five datasets.}
\label{label_format}
\end{table}

\begin{figure}[t]
    \centering
    \includegraphics[width=0.47\textwidth]{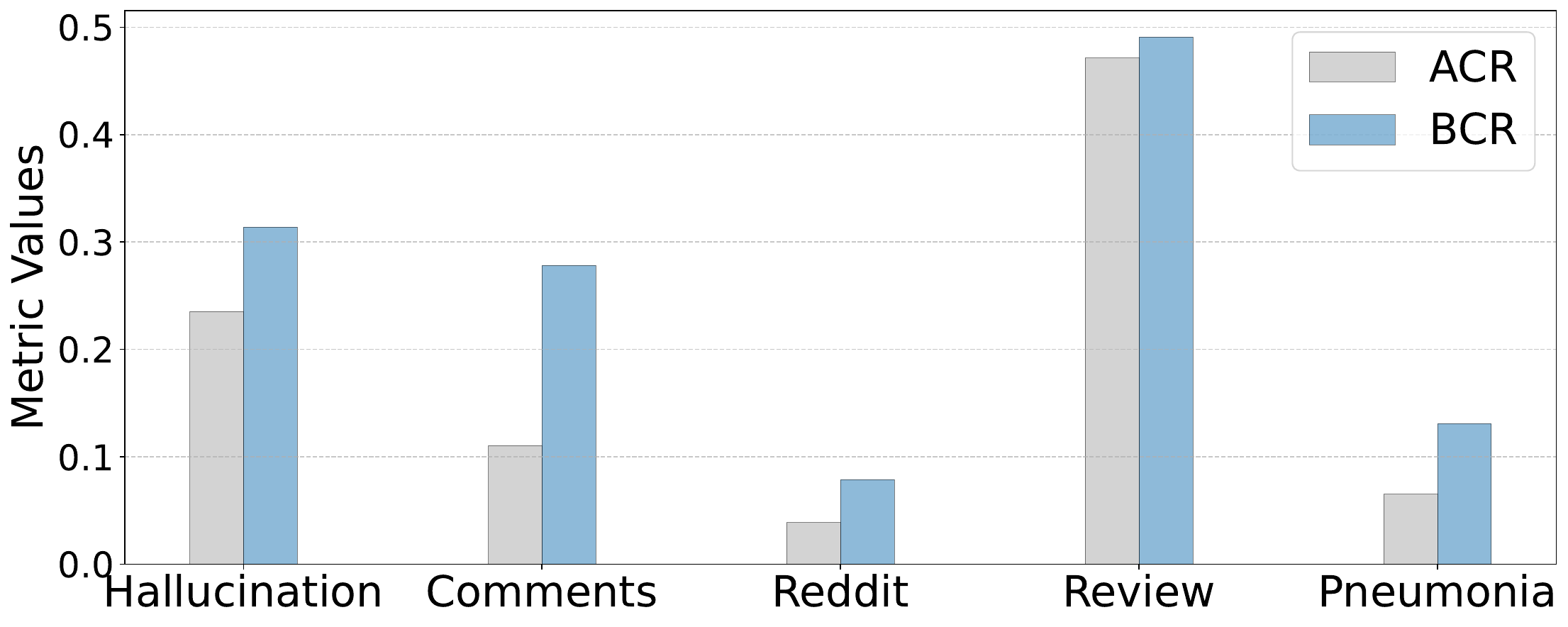}  
    \caption{Difference of predictions between correct label and flipped label demos: \emph{Adverse Correction Rate (ACR)} and \emph{Beneficial Correction Rate (BCR)} values under multiple hypotheses-based classification.}
    \label{difference}
\end{figure}
\section{Analysis}
\subsection{Is the result consistent with different in-context demonstration label formats?} 
To evaluate the consistency of results across different label formats, we compare two label formats: \emph{Label Format 1:} Demonstrations are provided as examples for positive and negative classes as in Figure \ref{prompt_template}.  \emph{Label Format 2:} Demonstrations are presented in the format of \emph{(Example, Label)}.

The average accuracy across all datasets for the correct and flipped label settings is presented in Figure \ref{label_format}. (Results for each dataset of \emph{Label Fomat 2} can be found in Appendix \ref{appendix:label_format_full}). With correct labels, the performance of the two label formats is very similar. However, in the flipped label settings, \emph{Label Format 2} shows almost no performance drop, which differs slightly from \emph{Label Format 1}. Notably, neither label format outperforms the hypotheses generated without demonstrations. This finding highlights the dominant role of the strong model prior, regardless of the presentation style of the demonstrations.
\begin{figure}[t]
    \centering
\includegraphics[width=0.47\textwidth]{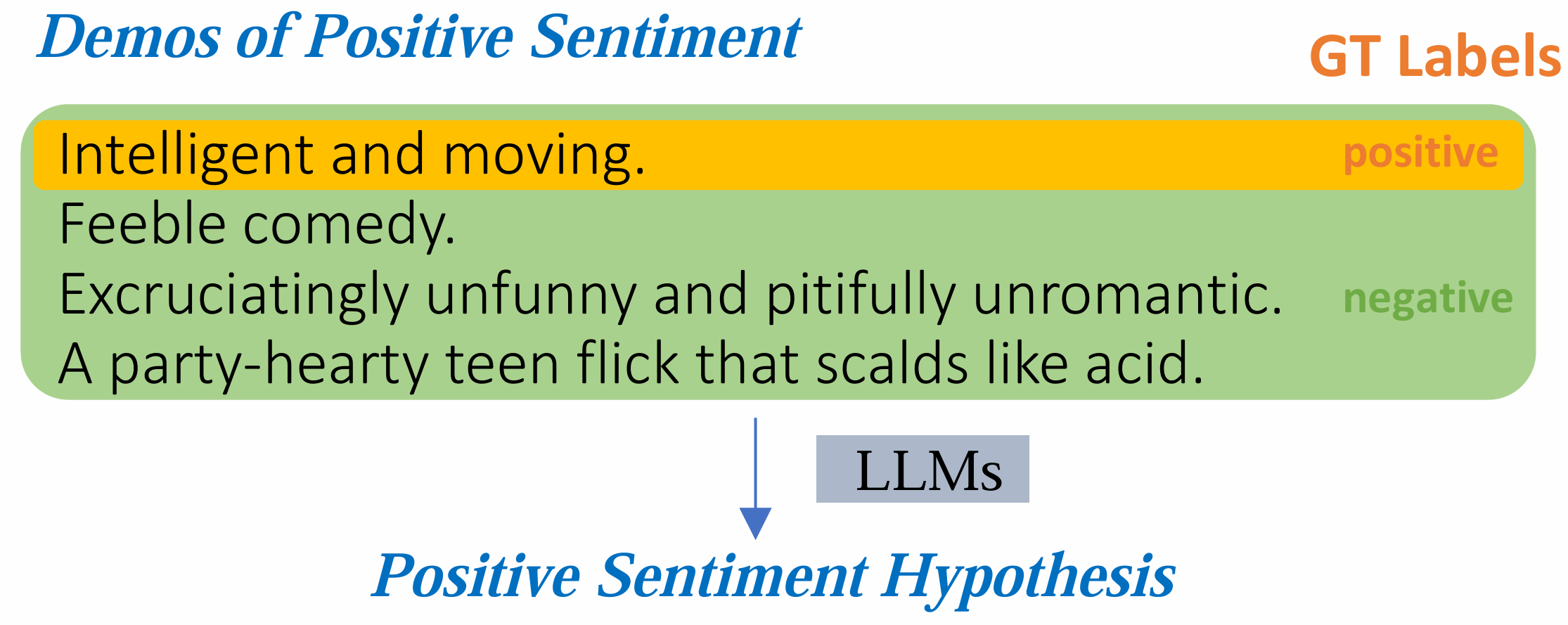}  
    \caption{An illustration of the case study: positive sentiment hypothesis generation. The highlighted text with a green background represents flipped label demos.}
    \label{case_study}
\end{figure}

\begin{figure}[t]
    \centering
\includegraphics[width=0.48\textwidth]{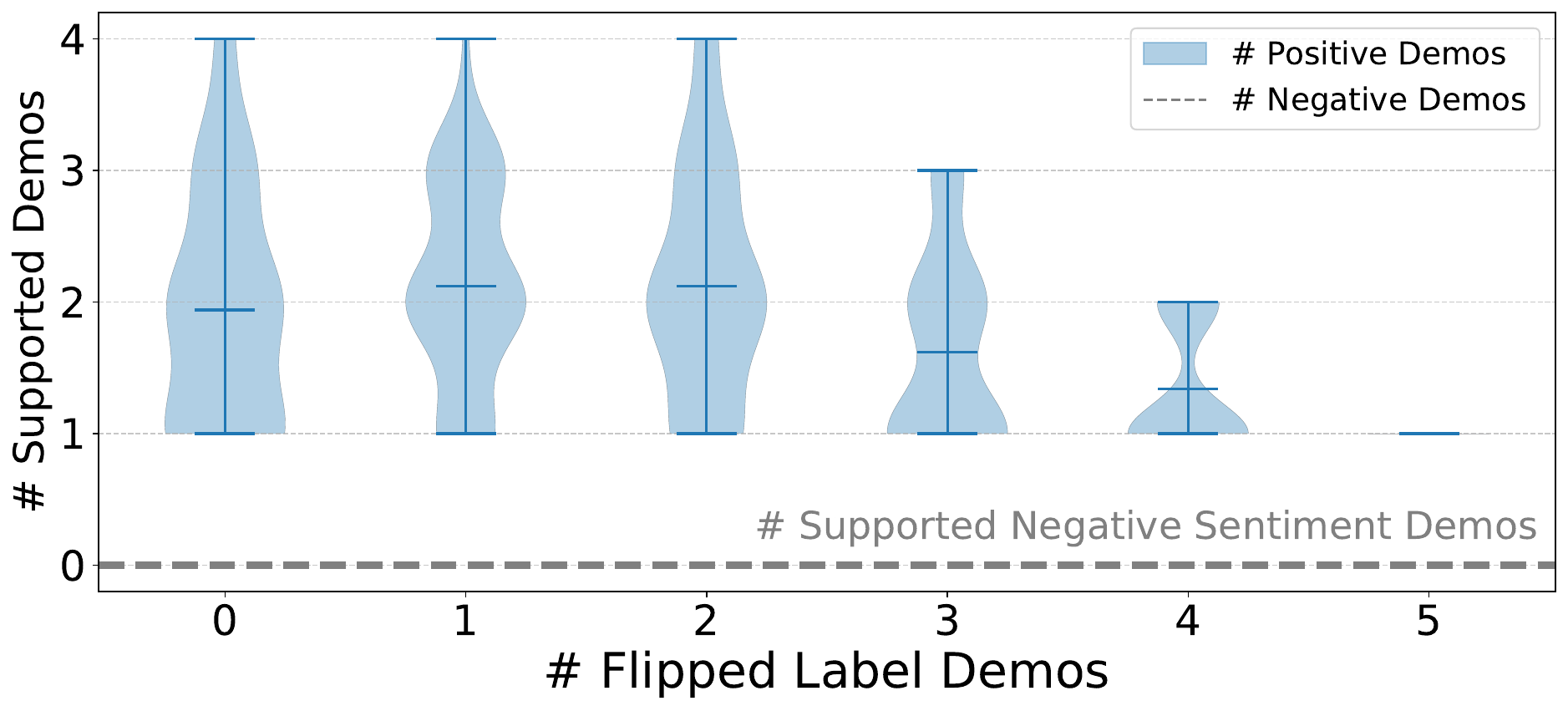}  
    \caption{Distribution of the number of supported true positive and negative demos with different number of flipped label demos. }
    \label{fig:sentiment}
\end{figure}
\subsection{What's the difference between correct label and flipped label settings?}
To get an deep understanding for the impact of flipping labels and provide a more fine-grained evaluation, we adopt two additional metrics introduced by \citet{wu2024strago}, Adverse Correction Rate (ACR) and Beneficial Correction Rate (BCR): 

{\small
\begin{equation}
\text{ACR} = \frac{\sum_{i=1}^{n} \mathbb{I}\left(y_{\text{correct}}(x_i) = y_i \wedge y_{\text{flipped}}(x_i) \neq y_i\right)}{\sum_{i=1}^{n} \mathbb{I}\left(y_{\text{correct}}(x_i) = y_i\right)},
\end{equation}
}

{\small
\begin{equation}
\text{BCR} = \frac{\sum_{i=1}^{n} \mathbb{I}\left(y_{\text{correct}}(x_i) \neq y_i \wedge y_{\text{flipped}}(x_i) = y_i\right)}{\sum_{i=1}^{n} \mathbb{I}\left(y_{\text{correct}}(x_i) \neq y_i\right)},
\end{equation}
}where $y_{\text{correct}}(x_i)$ and $y_{\text{flipped}}(x_i)$ represents the prediction results using the hypothesis generated with ground truth label and flipped label demonstrations, $x_i$, $y_i$ are input and ground truth label, respectively. These metrics offer a comprehensive evaluation of how flipping labels of the demonstrations influence the prediction results in downstream tasks. 

Results for multiple hypothesis-based classification prediction difference are shown in Table \ref{difference}. The results indicate that flipping the labels of in-context demonstrations does lead to some shifts in prediction outcomes, particularly notable in the truthful hotel review dataset, where nearly half of the predictions are affected. In contrast, for the other four datasets, label flipping only minimally alters prediction results. This suggests that while the model leverages the input-label mappings in provided demonstrations to inform its hypothesis generation, the inherent task-specific knowledge remains predominant, preventing the provided patterns from overriding its established priors.


\subsection{A Case Study: Hypothesis Generation for Positive Sentiment Pattern} This case study highlights that large language models (LLMs) heavily rely on prior knowledge when generating hypotheses, often ignoring patterns introduced in demonstrations. As shown in Figure \ref{case_study}, we replace true positive demonstrations with flipped label demonstrations (negative examples) to test whether the model adjusts its hypothesis or adheres to its prior.

Using IO-prompting, we provide six demonstrations, varying the number of flipped label demos from 0 to 5, and prompt the model to generate a hypothesis and corresponding supporting demonstrations. Repeating the experiment across 50 random seeds, we track the distribution of true positive and negative examples within the model's supported demonstrations for its hypothesis.

The results, shown in Figure \ref{fig:sentiment}, reveal notable patterns. The distribution of positive examples in the supported demonstrations begins to shift when three flipped label demonstrations are introduced. When five flipped demonstrations are provided, the mean number of positive examples converges to one. However, the model consistently avoids using flipped label demonstrations in its hypothesis generation, even when five demonstrations are flipped. 
This indicates that the model's hypotheses are predominantly influenced by prior knowledge rather than the provided demonstrations.


\section{Conclusion}
In this paper, we explore the role of task-specific priors in a real-world inductive reasoning scenario—hypothesis generation from labeled data. Experiments reveal that LLMs rely heavily on strong priors, which are difficult to override with demonstrations, offering insights into hypothesis generation mechanisms and future research directions.

\section*{Limitations}
\paragraph{Beyond Classification Problems.} Our experiments are limited to classification problems. Extensions to multi-choice or other tasks requires better representation of the hypothesis. We leave extensions to non-classification tasks for future work.
\paragraph{Better Application of Generated Hypotheses.} We think future can explore better application of generated hypotheses. For instance, this paper uses hypotheses to construct patterns for classification problems. Better application of hypotheses can improve downstream task performance, which we leave for future work.



\bibliography{anthology,custom}
\bibliographystyle{acl_natbib}

\appendix
\clearpage

\section{Dataset Details}
In this paper, we include 5 real-world datasets: hallucination, unhealthy comments in conversation, truthful hotel review, pneumonia MNIST and funny reddit post.

\begin{table}[ht]
    \centering
    \resizebox{0.5\textwidth}{!}{ 
        \begin{tabular}{lccc}
            \toprule
            \textbf{Dataset} & \textbf{Train} & \textbf{Validation} & \textbf{Test} \\
            \midrule
            Hallucination & 400 & 100 & 374 \\
            Pneumonia MNIST & 800 & 270 & 468 \\
            Unhealthy Conversation & 800 & 400 & 800 \\
            Funny Reddit & 200 & 100 & 308 \\
            Truthful Hotel Review & 800 & 300 & 500 \\
            \bottomrule
        \end{tabular}
        }
    \caption{Dataset Split for Train, Validation, and Test Sets.}
    \label{appendix：dataset_split}
\end{table}

\paragraph{Hallucination Pattern.} The dataset is first introduced in \citep{li2023evaluating}. We use its adversarial sampling version, which can be found in \href{https://github.com/RUCAIBox/POPE}{https://github.com/RUCAIBox/POPE}. To build our hallucination dataset, we prompt GPT-4o with each image-question pair once and see if the model hallucinates the object presence. As a result, we get 437 hallucinated image-question pairs and randomly sample another 437 image-question pairs as non-hallucination cases. 
\paragraph{Unhealthy Comments.}  Expert-annotated unhealthy conversations are from \citep{price2020six}, and we use the version from \citep{zhong2023goal}, which can be downloaded from \href{https://github.com/ruiqi-zhong/D5}{https://github.com/ruiqi-zhong/D5}. We sample longest 1000 samples for unhealthy and healthy comments from the dataset in our evaluation.
\paragraph{Truthful Hotel Reviews.} Truthful review detection is an instance of deception. The dataset we use is from \citep{zhou2024hypothesis}. The dataset includes 800 genuine reviews and 800 fictitious reviews for 20 hotels in Chicago, which can be downloaded from \href{https://github.com/ChicagoHAI/hypothesis-generation}{https://github.com/ChicagoHAI/hypothesis-generation}.
\paragraph{Funny Reddit Posts.} We collect jokes posted on the Reddit forum r/Jokes and cleaned by \citep{zhong2023goal}. This dataset can be downloaded from \href{https://github.com/ruiqi-zhong/D5}{https://github.com/ruiqi-zhong/D5}. We also remove all the duplicate samples for better dataset quality. 
\paragraph{Pneumonia MNIST.} Pneumonia recognition via chest X-ray image is an important problem. The dataset is from \citep{yang2023medmnist}, and can be downloaded from \href{https://medmnist.com/}{https://medmnist.com/}.

For each dataset, we have at least 200 samples for training, 100 samples for validation and 300 samples for test. For each dataset, we keep a balance between positive and negative class. Detailed statistics is shown in Table \ref{appendix：dataset_split}. 
\label{sec:dataset appendix}
\section{Implementation Details}
\paragraph{Model Parameters.}For API usage, the temperature and top-p are set to a small number $1 \times 10^{-15}$ and $1 \times 10^{-10}$, respectively.
\paragraph{Iterative Refinement. } We initialize the hypothesis bank with 5 hypotheses generated using IO-prompting. In refinement process, for each iteration, we select 5 hypotheses achieving highest accuracy on the validation set to LLMs for refinement and hope to get hypothesis with better quality. We evaluate 5 hypotheses with the best performance on validation dataset. We set refinement iteration to 3 in the paper. 

\paragraph{HypoGeniC. } We set the hypothesis bank size to $5$. Throughout the experiment, we use the reward efficient $\alpha = 0.5$, the number of initialized examples $num \_ init = 10$, and maximum number of wrong examples for each group to $2$ for more updates. For each iteration, we select top $3$ hypotheses to evaluate. For each update, we generate $1$ new hypothesis with incorrect examples. When there are no demonstrations, we rank the hypotheses in the bank by reward scores and use this ranking as feedback to get better hypothesis. 

\label{appendix: implementation details}
\section{Additional Results}
\subsection{Hypothesis-based Inference without task-specific knowledge}
\label{appendix:without knowledge}
To minimize the impact of prior knowledge in hypothesis-based inference, we eliminate task-specific knowledge from the evaluation prompt and remove learned patterns from the hypothesis. Instead, we reformulate the task into its corresponding modalities, prompting large language models (LLMs) with: \emph{"Does the provided text/image/image-question align with the given text/image/image-question patterns?"} This approach isolates the quality of the hypothesis, ensuring that inference is not influenced by prior knowledge.

The results are shown as Table \ref{without_prior}. On average, there is limited difference between the hypotheses generated with and without demonstrations. The findings demonstrate again that LLMs are able to generate hypothesis with high quality only with task-specific prior. 
\begin{table*}[ht]
\centering
\resizebox{0.90\textwidth}{!}{
\begin{tabular}{@{}llcccccc@{}}
\toprule
                  &    \textbf{Demos}            & \textbf{Hallucination} & \textbf{Unhealthy Comments} & \textbf{Funny Reddit} & \textbf{Truthful Review} & \textbf{PneumoniaMNIST} & \textbf{Overall Average} \\ \midrule
\multirow{2}{*}{Best} & w/o         &      63.1       & 70.1 & 61.6  & 64.0 & 75.6  & 66.9 \\
                      & w/          &      57.5      & 68.0 & 59.1   & 64.6 & 80.8 & 66.0 \\ \midrule
\multirow{2}{*}{Average} & w/o      &      54.4       & 60.3 & 54.1  & 56.7 & 69.8 & 59.1 \\
                         & w/       &      53.6       & 63.3  & 54.8 & 51.8 & 73.1 & 59.3 \\ \bottomrule
\end{tabular}
}
\caption{Accuraccy comparison of \textit{single hypothesis-based classification} without task-specific knowledge in inference: accuracy for the single hypothesis and the average across five hypotheses, with (\textbf{w/}) and without (\textbf{w/o}) demonstrations.}
\label{without_prior}
\end{table*}

\subsection{Results of Different Datasets with Label Format 2}
We provide results on each dataset with \emph{Label Format 2}. The results are shown as Table \ref{labelformat2}. From the results, we can see that the results vary by dataset. However, there is quite limited difference (smaller than $3\% $) between correct and flipped label settings, showing the prior is too strong to be overridden by provided demonstrations.
\label{appendix:label_format_full}
\begin{table*}[ht!]
\centering
\resizebox{0.9\textwidth}{!}{
\renewcommand{\arraystretch}{1.2}
\setlength{\tabcolsep}{6pt} 
\begin{tabular}{lcccccc}
\toprule
\textbf{Label} & \textbf{Hallucination} & \textbf{Unhealthy Comments} & \textbf{Funny Reddit} & \textbf{Truthful Review} & \textbf{PneumoniaMNIST} & \textbf{Average} \\
\midrule
Correct (Best) & 63.9 & 70.6 & 61.7 & 68.0 & 75.2 & 67.9  \\
Flipped (Best) & 61.2 & 71.5 & 62.0 & 68.8 & 73.9 & 67.5 \\
\midrule
Correct (Avg)  & 57.0 & 65.1 & 59.0 & 62.5 & 70.3 & 62.8  \\
Flipped (Avg)  & 57.8 & 64.7 & 57.3 & 61.3 & 68.5 & 61.9  \\
\bottomrule
\end{tabular}
}
\caption{Accuracy comparison across five datasets with correct and flipped label settings in the \emph{Label Format 2}.}
\label{labelformat2}
\end{table*}

\subsection{Results with Gemini Model}
\label{appendix:gemini}
We test IO-prompting with and without demonstrations on model \textbf{gemini-1.5-pro-002}. We report the average over two random seeds. The results are shown as Table \ref{gemini}. On average, there is quite limited performance difference with and without demonstrations, demonstrating that with only prior, LLMs can generate good hypotheses. 
\begin{table*}[ht]
\centering
\resizebox{0.9\textwidth}{!}{
\begin{tabular}{@{}llcccccc@{}}
\toprule
                  &    \textbf{Demos}            & \textbf{Hallucination} & \textbf{Unhealthy Comments} & \textbf{Funny Reddit} & \textbf{Truthful Review} & \textbf{PneumoniaMNIST} & \textbf{Overall Average} \\ \midrule
\multirow{2}{*}{Best} & w/o         &      -       & 67.9 $\pm$ 0.2 & 62.7 $\pm$ 0.3 & 68.8 $\pm$ 2.0 & 58.2 $\pm$ 1.8 & 64.4 \\
                      & w/          &      -       & 67.8 $\pm$ 1.3 & 65.9 $\pm$ 0.3 & 66.9 $\pm$ 1.7 & 55.7 $\pm$ 1.4 & 64.1 \\ \midrule
\multirow{2}{*}{Average} & w/o      &       -      & 61.9 $\pm$ 0.3 & 56.8 $\pm$ 0.0 & 64.5 $\pm$ 2.3 & 53.1 $\pm$ 0.2 & 59.1 \\
                         & w/       &      -       & 62.4 $\pm$ 1.1 & 58.0 $\pm$ 1.2 & 63.4 $\pm$ 1.2 & 53.0 $\pm$ 1.5 & 59.2 \\ \bottomrule
\end{tabular}
}
\caption{Accuraccy comparison of \textit{single hypothesis-based classification} with \textbf{gemini-1.5-pro-002}: accuracy (\textit{mean} ± \textit{standard deviation}) for the best single hypothesis and the average across five hypotheses, with (\textbf{w/}) and without (\textbf{w/o}) demonstrations. "-" means the response is prohibited due to satety reasons. }
\label{gemini}
\end{table*}
\section{Evaluation Details}
\paragraph{LLM-based Evaluation Details.}
We prompt large language models (LLMs) to generate five hypotheses for each dataset across three different baselines. This results in a total of 25 hypotheses per baseline for both settings: with and without demonstrations.

For LLM-based scoring, each hypothesis is evaluated by prompting the LLMs to assign a score on a 1–5 scale. Additionally, for pairwise comparisons, we randomly pair hypotheses generated with and without demonstrations, creating a total of 25 pairs for evaluation.
\paragraph{Human Evaluation Details.}
We randomly pair the hypotheses generated with and without demonstrations across three datasets and three baselines. We selected the datasets unhealthy comments, truthful reviews, and funny Reddit posts because their domain knowledge is accessible to non-experts.

\begin{figure*}[t]
    \centering
    \includegraphics[width=0.9\textwidth]{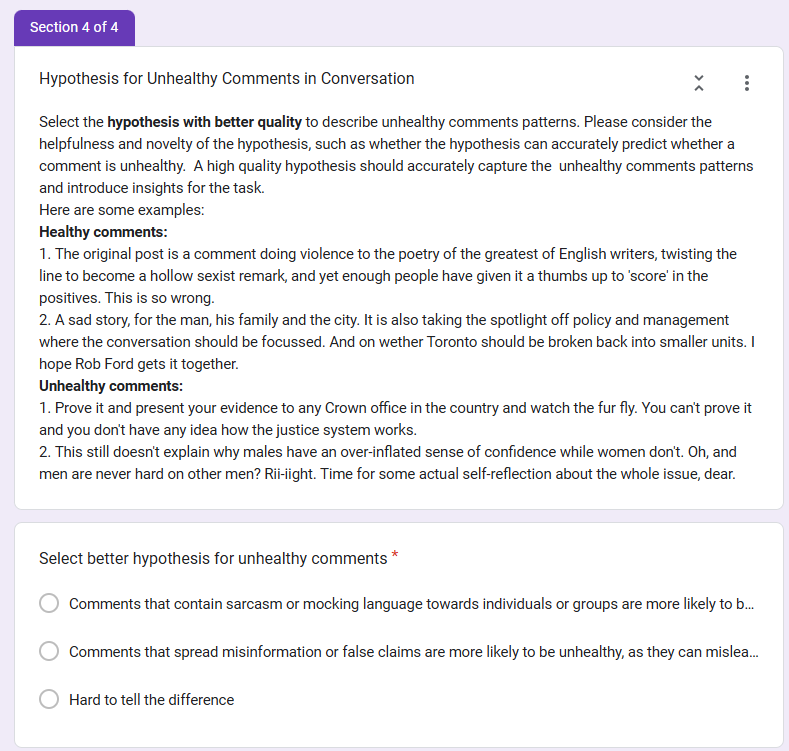}  
    \caption{Example interface of human evaluation.}
    \label{example_human_interface}
\end{figure*}

Participants were provided with a questionnaire for evaluation. For each evaluation, we included the evaluation context, paired hypotheses, and illustrative examples to guide participants. An example of the evaluation interface is shown in Figure \ref{example_human_interface}.
\label{appendix: qualitative study}
\section{Examples of Generated Hypothesis}
We randomly select generated hypothesis with and without demonstrations for each dataset, shown as Table \ref{hypothesis_examples}.
\label{appendix: example_hypothesis}
\begin{table*}[ht!]
    \centering
    \resizebox{0.9\textwidth}{!}{
    \begin{tabular}{l p{6cm} p{6cm}}
        \toprule
        \textbf{Dataset} & \textbf{Hypothesis without Demos} & \textbf{Hypothesis with Demos} \\
        \midrule
        Hallucination& 
        Hallucinations are more likely to occur when the questioned object is partially occluded or located in a cluttered environment, making it difficult for the model to accurately identify its presence or absence. &
         **Complex Backgrounds Hypothesis**: Images with complex or cluttered backgrounds may lead to hallucinations, as the model might misinterpret overlapping or densely packed objects as the queried item. \\
        
        \midrule
        Unhealthy Comments & 
        Comments containing personal attacks or insults are more likely to be unhealthy, as they often escalate conflicts and discourage constructive dialogue. & 
        Comments that include personal attacks or derogatory language towards individuals are more likely to be unhealthy.  \\
        
        \midrule
        Funny Reddit Posts & 
        Posts that incorporate unexpected punchlines or twists are more likely to be perceived as funny, as they play on the element of surprise and subvert reader expectations. & 
        Posts that use wordplay or double entendres, where a phrase can be interpreted in multiple humorous ways, tend to be perceived as funny. \\
        
        \midrule
        Pneumonia MNIST & 
        The presence of pleural effusion, seen as blunting of the costophrenic angles or fluid layering in the pleural space, may indicate pneumonia. & 
        Presence of air bronchograms within areas of increased opacity suggests pneumonia.\\
        
        \midrule
        Truthful Hotel Reviews & 
        Truthful reviews often mention both positive and negative aspects of the stay, providing a balanced perspective rather than an overly positive or negative one.& 
        Truthful reviews often mention both positive and negative aspects of the stay, providing a balanced perspective that suggests authenticity.\\
        
        \bottomrule
    \end{tabular}
    }
    \caption{Examples of Generated Hypotheses with and without In-Context Demonstrations.}
    \label{hypothesis_examples}
\end{table*}

\section{Prompts}
For prompt construction, we begin by manually crafting a prompt for hallucination pattern induction, following a format similar to that used in \citep{zhou2024hypothesis}. Subsequently, we leverage in-context learning to generate prompts for other tasks. Specifically, we provide the task name along with the manually constructed prompt to the language model, enabling it to generate prompts tailored to other tasks.

\label{appendix:prompt}

\begin{figure*}[t]
\begin{tcolorbox}[colback=orange!5!white,colframe=orange!75!black,title= Prompt for hallucination with demonstrations,width=\textwidth]
You're a professional vision-language model behavior analyst.\\
Given a set of image-question pairs, we want to generate hypotheses that are useful for predicting whether a model will hallucinate the existence of an object in response to a given question. \\
In other words, we want to know whether the model will falsely claim the presence of an object in the image when answering the question. \\
Using the given examples, please propose \{\{num\_hypotheses\}\} possible hypotheses that can identify specific patterns that occur across the provided image-question pairs. \\
Each hypothesis should contain the following: a hypothesis about what image content features, object features, or contextual relationships make the model more likely to hallucinate. \\
The hypotheses should analyze what kinds of image-question pairs are more likely to trigger hallucinations.\\

Some examples of hallucination and non-hallucination cases are shown.\\
Hallucination cases are from number 1 to \{\{num\_1\}\}, and non-hallucination cases are from number \{\{num\_2\}\} to \{\{num\_3\}\}.\\

Based on provided examples, please generate hypotheses that are useful for predicting whether the model will hallucinate the existence of an object in response to a given question. \\
Propose \{\{num\_hypotheses\}\} possible hypotheses for hallucination patterns. \\
Generate them in the format of 1. [hypothesis], 2. [hypothesis], ... \{\{num\_hypotheses\}\}. [hypothesis].\\
Proposed hypotheses:
\end{tcolorbox}
\end{figure*}

\begin{figure*}[t]
\begin{tcolorbox}[colback=orange!5!white,colframe=orange!75!black,title=Prompt for hallucination without demonstrations,width=\textwidth]
You are an expert in vision-language models, specializing in detecting and preventing hallucinations.\\
We want to generate hypotheses that are useful for predicting whether a vision-language model will hallucinate the existence of an object when responding to a question about an image. \\
In other words, we want to identify patterns that indicate when the model will incorrectly claim the presence of an object not present in the image, or the absence of an object that is present.\\
Please propose \{\{num\_hypotheses\}\} possible hypotheses.\\
These hypotheses should identify specific patterns that occur across common hallucination cases and focus on the relationship between the image content and the questioned object.\\
Each hypothesis should contain the following: a hypothesis about what image content features, object features, or contextual relationships make the model more likely to hallucinate.\\
The hypotheses should analyze what kind of image-question pairs are more likely to lead to hallucinations.\\
Please generate \{\{num\_hypotheses\}\} possible hypotheses for hallucination patterns in the given context.\\
Generate them in the format of 1. [hypothesis], 2. [hypothesis], ... \{\{num\_hypotheses\}\}. [hypothesis].\\
Don't talk about any other words.\\
Proposed hypotheses:
\end{tcolorbox}
\end{figure*}
\begin{figure*}[t]
\begin{tcolorbox}[colback=orange!5!white,colframe=orange!75!black,title=Prompt for unhealthy comments with demonstrations,width=\textwidth]
You're an expert comment analyst in online conversation.\\
Given a set of comments, we want to generate hypotheses that are useful for predicting whether a comment is unhealthy. \\
In other words, we want to know if the comment contributes to unhealthy conversations online.\\
Using the given examples, please propose \{\{num\_hypotheses\}\} possible hypotheses.\\
These hypotheses should identify specific patterns that occur across the provided unhealthy comments.
Each hypothesis should contain the following: A hypothesis about what makes comments more likely to be unhealthy.
The hypotheses should analyze what kind of comments are likely to be unhealthy.\\
Here are some examples of unhealthy and healthy comments:\\

Unhealthy comments:\\
\{\{positive\_examples\}\}\\
Healthy comments:\\
\{\{negative\_examples\}\}\\

Based on the provided examples, please generate hypotheses that are useful for predicting whether a comment is unhealthy.\\
Propose \{\{num\_hypotheses\}\} possible hypotheses for unhealthy comment patterns.\\
Generate them in the format of 1. [hypothesis], 2. [hypothesis], ... \{\{num\_hypotheses\}\}. [hypothesis].\\
Don't include any other words.\\
Proposed hypotheses:
\end{tcolorbox}
\end{figure*}
\begin{figure*}[t]
\begin{tcolorbox}[colback=orange!5!white,colframe=orange!75!black,title=Prompt for unhealthy comments without demonstrations,width=\textwidth]
You're an expert comment analyst in online conversation.\\
We want to generate hypotheses that are useful for predicting whether a comment is unhealthy. In other words, we want to know if the comment contributes to unhealthy conversations online.\\
Please propose \{\{num\_hypotheses\}\} possible hypotheses.\\
These hypotheses should identify specific patterns that occur across common unhealthy comments.\\
Each hypothesis should contain the following: A hypothesis about what makes comments more likely to be unhealthy.\\
The hypotheses should analyze what kind of comments are likely to be unhealthy.\\
Please generate hypotheses that are useful for predicting whether a comment is unhealthy or healthy.\\
Propose \{\{num\_hypotheses\}\} possible hypotheses for unhealthy comment patterns.\\
Generate them in the format of 1. [hypothesis], 2. [hypothesis], ... \{\{num\_hypotheses\}\}. [hypothesis].\\
Don't talk about any other words.\\
Proposed hypotheses:
\end{tcolorbox}
\end{figure*}
\begin{figure*}[t]
\begin{tcolorbox}[colback=orange!5!white,colframe=orange!75!black,title=Prompt for truthful reviews with demonstrations,width=\textwidth]
You're a professional hotel review analyst.\\
Given a set of hotel reviews, we want to generate hypotheses that are useful for predicting whether a review is truthful. In other words, we want to know whether the review is written by someone who actually lived in the hotel.\\
Using the given examples, please propose \{\{num\_hypotheses\}\} possible hypotheses.\\
These hypotheses should identify specific patterns that occur across the provided reviews.
Each hypothesis should contain the following: A hypothesis about what makes reviews more likely to be truthful.
The hypotheses should analyze what kind of reviews are likely to be truthful.\\
Here are some examples of truthful and deceptive reviews:\\

Truthful reviews:\\
\{\{positive\_examples\}\}\\
Deceptive reviews:\\
\{\{negative\_examples\}\}\\

Based on provided examples, please generate hypotheses that are useful for predicting whether a review is truthful.\\
Propose \{\{num\_hypotheses\}\} possible hypotheses for truthful review patterns.\\
Generate them in the format of 1. [hypothesis], 2. [hypothesis], ... \{\{num\_hypotheses\}\}. [hypothesis].\\
Don't talk about any other words.\\
Proposed hypotheses:
\end{tcolorbox}
\end{figure*}

\begin{figure*}[t]
\begin{tcolorbox}[colback=orange!5!white,colframe=orange!75!black,title=Prompt for truthful reviews with demonstrations,width=\textwidth]
You're a professional hotel review analyst.\\
We want to generate hypotheses that are useful for predicting whether a review is truthful or deceptive. In other words, we want to know whether the review is written by someone who actually lived in the hotel.\\
Please propose \{\{num\_hypotheses\}\} possible hypotheses.\\
These hypotheses should identify specific patterns that occur across common truthful reviews.
Each hypothesis should contain the following: A hypothesis about what makes reviews more likely to be truthful.
The hypotheses should analyze what kind of reviews are likely to be truthful or deceptive.\\
Please generate hypotheses that are useful for predicting whether a review is truthful or deceptive.\\
Propose \{\{num\_hypotheses\}\} possible hypotheses for truthful review patterns.\\
Generate them in the format of 1. [hypothesis], 2. [hypothesis], ... \{\{num\_hypotheses\}\}. [hypothesis].\\
Don't talk about any other words.\\
Proposed hypotheses:
\end{tcolorbox}
\end{figure*}
\begin{figure*}[t]
\begin{tcolorbox}[colback=orange!5!white,colframe=orange!75!black,title=Prompt for PneumoniaMNIST with demonstrations,width=\textwidth]
You're a professional radiologist specializing in chest X-rays.\\
Given a set of labeled chest X-ray images, we want to generate hypotheses that are useful for predicting whether a patient has pneumonia. In other words, we want to know whether the X-ray shows signs of pneumonia.\\
Using the given examples, please propose \{\{num\_hypotheses\}\} possible hypotheses.\\
These hypotheses should identify specific patterns that occur across the provided X-ray images.\\
Each hypothesis should contain the following: A hypothesis about what makes an X-ray more likely to indicate pneumonia.
The hypotheses should analyze what kind of image patterns are likely to be indicative of pneumonia or not.\\

Some examples of X-ray images labeled as pneumonia and non-pneumonia are shown.\\
Pneumonia images are from number 1 to \{\{num\_1\}\}, and non-pneumonia images are from number \{\{num\_2\}\} to \{\{num\_3\}\}.\\

Based on provided examples, please generate hypotheses that are useful for predicting whether an X-ray shows pneumonia or not.\\
Propose \{\{num\_hypotheses\}\} possible hypotheses for pneumonia pattern recognition.\\
Generate them in the format of 1. [hypothesis], 2. [hypothesis], ... \{\{num\_hypotheses\}\}. [hypothesis].\\
Don't include any other information.\\
Proposed hypotheses:
\end{tcolorbox}
\end{figure*}
\begin{figure*}[t]
\begin{tcolorbox}[colback=orange!5!white,colframe=orange!75!black,title=Prompt for PneumoniaMNIST without demonstrations,width=\textwidth]
You're a professional radiologist.\\
We want to generate hypotheses that are useful for predicting whether a patient has pneumonia based on their chest X-ray image. In other words, we want to know which patterns in the image are indicative of pneumonia presence.\\
Please propose \{\{num\_hypotheses\}\} possible hypotheses.\\
These hypotheses should identify specific visual patterns that occur in typical pneumonia cases.\\
Each hypothesis should contain the following: A hypothesis about what makes an image more likely to show signs of pneumonia.\\
The hypotheses should analyze what kind of visual patterns or markers are likely to indicate pneumonia.\\
Please generate hypotheses that are useful for predicting whether a patient has pneumonia or not based on the X-ray.\\
Propose \{\{num\_hypotheses\}\} possible hypotheses for pneumonia-related visual patterns.\\
Generate them in the format of 1. [hypothesis], 2. [hypothesis], ... \{\{num\_hypotheses\}\}. [hypothesis].\\
Don't include any additional context.\\
Proposed hypotheses:
\end{tcolorbox}
\end{figure*}
\begin{figure*}[t]
\begin{tcolorbox}[colback=orange!5!white,colframe=orange!75!black,title=Prompt for funny reddit with demonstrations,width=\textwidth]
You're a professional humor analyst for Reddit posts.\\
Given a set of Reddit posts, we want to generate hypotheses that are useful for predicting whether a post is considered funny or not. In other words, we want to know whether a post contains humor patterns often associated with successful humorous posts.\\
Using the provided examples, please propose \{\{num\_hypotheses\}\} possible hypotheses.\\
These hypotheses should identify specific patterns that occur across the provided posts.\\
Each hypothesis should contain the following: A hypothesis about what makes posts more likely to be considered funny.
The hypotheses should analyze what kind of posts are likely to be perceived as funny or not.\\
Here are some examples of funny and unfunny posts:\\

Funny posts:\\
\{\{positive\_examples\}\}\\
Unfunny posts:\\
\{\{negative\_examples\}\}\\

Based on the provided examples, please generate hypotheses that are useful for predicting whether a post is funny or not.\\
Propose \{\{num\_hypotheses\}\} possible hypotheses for funny post patterns.\\
Generate them in the format of 1. [hypothesis], 2. [hypothesis], ... \{\{num\_hypotheses\}\}. [hypothesis].\\
Don't talk about any other words.\\
Proposed hypotheses:
\end{tcolorbox}
\end{figure*}
\begin{figure*}[t]
\begin{tcolorbox}[colback=orange!5!white,colframe=orange!75!black,title=Prompt for funny reddit without demonstrations,width=\textwidth]
You're a professional Reddit content analyst.\\
We want to generate hypotheses that are useful for predicting whether a Reddit post is funny or not.\\
In other words, we want to know what characteristics make a post likely to be perceived as humorous by the community.\\
Please propose \{\{num\_hypotheses\}\} possible hypotheses.\\
These hypotheses should identify specific patterns that occur across common funny posts.\\
Each hypothesis should contain the following: A hypothesis about what makes posts more likely to be perceived as funny.\\
The hypotheses should analyze what kind of posts are likely to be considered humorous or non-humorous.\\
Please generate hypotheses that are useful for predicting whether a post is funny or not.\\
Propose \{\{num\_hypotheses\}\} possible hypotheses for funny Reddit post patterns.\\
Generate them in the format of 1. [hypothesis], 2. [hypothesis], ... \{\{num\_hypotheses\}\}. [hypothesis].\\
Don't talk about any other words.\\
Proposed hypotheses:
\end{tcolorbox}
\end{figure*}
\begin{figure*}[t]
\begin{tcolorbox}[colback=orange!5!white,colframe=orange!75!black,title=Evaluation prompt for hallucination,width=\textwidth]
You are an expert in vision-language model behavior detection, and your job is to apply learned patterns to predict if the model will hallucinate for the given image and question.\\
Here are some previously learned hallucination patterns:\\
\{\{patterns\}\}\\
The image is shown and the question is: \{\{text\}\}\\
Based on the learned patterns, will the model hallucinate?\\
Let's think step by step.\\
Step 1: Look at the given image and question, and compare them with the provided hallucination patterns.\\
Step 2: If the image and question pair matches learned hallucination pattern, then it is likely that the model will hallucinate. If it does not match the pattern, then it is likely that the model will not hallucinate.\\
Step 3: Apply the pattern to the given image-question pair and predict whether the model will hallucinate based on the learned patterns.\\
Step 4: Give your final answer: yes or no. If you are unsure, respond with no.\\
Please give your answer strictly in the following format:\\
\texttt{"""\\
Analysis: [your step-by-step analysis]\\
Answer: [your answer]\\
"""\\}
Give your answer in the above format. Don't talk about any other words.
\end{tcolorbox}
\end{figure*}
\begin{figure*}[t]
\begin{tcolorbox}[colback=orange!5!white,colframe=orange!75!black,title=Evaluation prompt for unhealthy comments,width=\textwidth]
You will predict whether a comment is unhealthy based on the provided unhealthy comment patterns.\\
Here are some previously generated unhealthy comment patterns:\\
\{\{patterns\}\}\\
A comment is the following: \{\{text\}\}\\
Is this comment unhealthy?\\
Think step-by-step.\\
Step 1: Look at the new comment and compare it with the provided unhealthy comment patterns.\\
Step 2: If the comment matches the pattern, then it is likely unhealthy. If it does not match the pattern, then it is likely healthy.\\
Step 3: Apply the pattern to the new comment and predict whether the new comment is unhealthy.\\
Step 4: Give your final answer: yes or no. If you are unsure, respond with no.\\
Please give your answer strictly in the following format:\\
\texttt{"""\\
Analysis: [your step-by-step analysis]\\
Answer: [your answer]\\
"""\\}
\end{tcolorbox}
\end{figure*}
\begin{figure*}[t]
\begin{tcolorbox}[colback=orange!5!white,colframe=orange!75!black,title=Evaluation prompt for truthful review,width=\textwidth]
You will predict whether a hotel review is truthful based on the given truthful review patterns.\\
Here are some previously generated truthful review patterns:\\
\{\{patterns\}\}\\
A hotel review is the following: \{\{text\}\}\\
Is this hotel review truthful?\\
Think step-by-step.\\
Step 1: Look at the new hotel review and compare it with the provided truthful review patterns.\\
Step 2: If the review matches the pattern, then it is likely truthful. If it does not match the pattern, then it is likely not truthful.\\
Step 3: Apply the pattern to the new hotel review and predict whether the new hotel review is truthful.\\
Step 4: Give your final answer: yes or no. If you are unsure, respond with no.\\
Please give your answer strictly in the following format:\\
\texttt{"""\\
Analysis: [your step-by-step analysis]\\
Answer: [your answer]\\
"""\\}
\end{tcolorbox}
\end{figure*}
\begin{figure*}[t]
\begin{tcolorbox}[colback=orange!5!white,colframe=orange!75!black,title=Evaluation prompt for pneumoniaMNIST,width=\textwidth]
You are an expert in pneumonia detection, and your job is to apply learned patterns to predict if a person has pneumonia.\\
Here are some previously generated pneumonia patterns: \{\{patterns\}\}\\
A chest X-ray image is shown.\\
Based on the learned patterns and given image, is this person likely to have pneumonia based on the learned patterns?\\
Think step-by-step.\\
Step 1: Look at the given chest X-ray image and compare it with the provided pneumonia patterns.\\
Step 2: If the image features match the pneumonia patterns, then the person is likely to have pneumonia. If the features do not match the patterns, then the person is likely not to have pneumonia.\\
Step 3: Apply the pattern to the new chest X-ray image and predict whether the person has pneumonia.\\
Step 4: Give your final answer: yes or no. If you are unsure, respond with no.\\
Please give your answer strictly in the following format:\\
\texttt{"""\\
Analysis: [your step-by-step analysis]\\
Answer: [your answer]\\
"""\\}
Give your answer in the above format. Don't talk about any other words.
\end{tcolorbox}
\end{figure*}
\begin{figure*}[t]
\begin{tcolorbox}[colback=orange!5!white,colframe=orange!75!black,title=Evaluation prompt for funny reddit,width=\textwidth]
You will predict whether a Reddit post is funny based on the given funny Reddit post patterns.\\
Here are some previously generated funny Reddit post patterns:\\
\{\{patterns\}\}\\
A Reddit post is the following: \{\{text\}\}\\
Is this Reddit post funny?\\
Think step-by-step:\\
Step 1: Look at the new Reddit post and compare it with the provided funny post patterns.\\
Step 2: If the post matches the pattern, then it is likely funny. If it does not match the pattern, then it is likely not funny.\\
Step 3: Apply the pattern to the new Reddit post and predict whether the new post is funny.\\
Step 4: Give your final answer: yes or no. If you are unsure, respond with no.\\
Please give your answer strictly in the following format:\\
\texttt{"""\\
Analysis: [your step-by-step analysis]\\
Answer: [your answer]\\
"""\\}
\end{tcolorbox}
\end{figure*}

\end{document}